\documentclass[10pt,twocolumn,letterpaper]{article}
\usepackage[accsupp]{axessibility} 
\usepackage[table,dvipsnames]{xcolor}
\usepackage{bm}
\usepackage{epsfig}
\usepackage{graphicx}
\usepackage{amsmath}
\newcounter{RNum}
\usepackage{booktabs}

\usepackage{multirow}
\usepackage{graphics} 
\usepackage{epsfig} 

\usepackage{cvpr}              

\usepackage{graphicx}
\usepackage{amsmath}
\usepackage{amssymb}
\usepackage{booktabs}

\usepackage{amsfonts}

\usepackage{graphicx}
\usepackage{pifont}
\usepackage{lipsum}
\usepackage[pagebackref,breaklinks,colorlinks]{hyperref}

\usepackage{float}
\usepackage{times}
\usepackage{mathrsfs}
\usepackage{epsfig}
\usepackage{graphicx}
\usepackage{graphics} 
\usepackage{tablefootnote}
\usepackage{graphicx}
\usepackage{grffile} 
\usepackage{amsmath} 
\usepackage{amssymb}  
\usepackage{color}
\usepackage{booktabs}
\usepackage{bm}
\usepackage{url}

\usepackage[noend]{algpseudocode}
\usepackage{algorithm}
\usepackage{makecell}
\usepackage[square, comma, numbers,sort&compress]{natbib}
\usepackage{multirow}

\usepackage{enumitem}
\usepackage[table,dvipsnames]{xcolor}

\usepackage{hyperref}
\hypersetup{
linkcolor=BrickRed
,citecolor=Green
,filecolor=Mulberry
,urlcolor=NavyBlue
,menucolor=BrickRed
,runcolor=Mulberry
,linkbordercolor=BrickRed
,citebordercolor=Green
,filebordercolor=Mulberry
,urlbordercolor=NavyBlue
,menubordercolor=BrickRed
,runbordercolor=Mulberry
}

\usepackage{bbding}
\usepackage{pifont}
\usepackage{wasysym}
\newcommand{\cmark}{\ding{51}}
\newcommand{\xmark}{\ding{55}}

\usepackage[capitalize]{cleveref}
\crefname{section}{Sec.}{Secs.}
\Crefname{section}{Section}{Sections}
\Crefname{table}{Table}{Tables}
\crefname{table}{Tab.}{Tabs.}

\def\cvprPaperID{9778} 
\def\confName{CVPR}
\def\confYear{2022}

\begin{document}

\title{Egocentric Prediction of Action Target in 3D}
\def\ourname{EgoPAT3D}
\author{Yiming Li$^{1,}\thanks{Equal contribution}$,\, Ziang Cao$^{2,}\footnotemark[1]$,\,   Andrew Liang$^1$,   {Benjamin Liang}$^1$, 
 {Luoyao Chen}$^1$,  {Hang Zhao}$^{3}$, {Chen Feng}$^{1,}$\thanks{The corresponding author is Chen Feng {\tt\small cfeng@nyu.edu}}\\
$^{1}$New York University \quad $^{2}$Tongji University \quad $^{3}$Tsinghua University
\\
{\tt\small \url{https://ai4ce.github.io/EgoPAT3D/}}
}

\twocolumn[{%
\maketitle
\vspace{-14mm}
\begin{figure}[H]
\begin{center}
\hsize=\textwidth 
\includegraphics[width=0.85\textwidth]{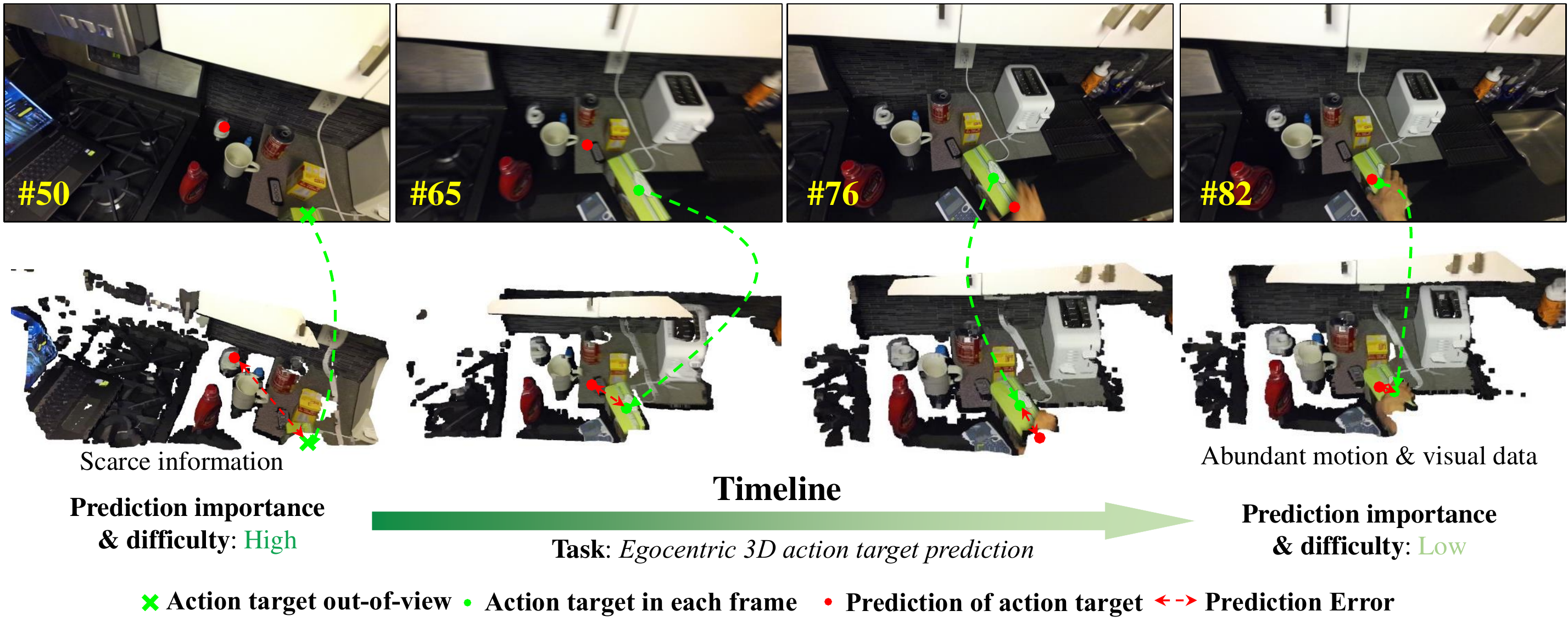}
\vspace{-8pt}
\caption{\textbf{Illustration of the proposed egocentric prediction task}. The \textcolor[rgb]{ 1,  0,  0}{predicted} and \textcolor{Green}{ground truth} target locations of an action (grasping) sequence are visualized by \textcolor[rgb]{ 1,  0,  0}{red} and \textcolor{Green}{green} dots. In the early stage, only scarce information is available, thus it is more challenging to achieve reliable prediction. Over time, more visual and motion cues are accumulated, making it easier to predict the action target. Note that the prediction is updated in each frame and the 3D action target in each frame is 
expressed in the coordinate system of that frame.}
\label{fig:task}
\end{center} 
\end{figure}
}]
\vspace{-1.5cm}
{
  \renewcommand{\thefootnote}%
    {\fnsymbol{footnote}}
  \footnotetext[1]{Equal contribution}
  \footnotetext[2]{The corresponding author is Chen Feng. {\tt\small cfeng@nyu.edu}}
}

\begin{abstract}
\vspace{-0.3cm}
We are interested in anticipating as early as possible the target location of a person's object manipulation action in a 3D workspace from egocentric vision. It is important in fields like human-robot collaboration, but has not yet received enough attention from vision and learning communities. To stimulate more research on this challenging egocentric vision task, we propose a large multimodality dataset of more than 1 million frames of RGB-D and IMU streams, and provide evaluation metrics based on our high-quality 2D and 3D labels from semi-automatic annotation. Meanwhile, we design baseline methods using recurrent neural networks and conduct various ablation studies to validate their effectiveness. Our results demonstrate that this new task is worthy of further study by researchers in robotics, vision, and learning communities.
\end{abstract}
\vspace{-0.4cm}

\section{Introduction}\label{sec:intro}
\vspace{-2mm}
Egocentric vision, which parses images from a wearable camera capturing a person's visual field, has been an important area in robotics and computer vision due to its wide applications, \textit{e.g.}, virtual reality~\citep{Jang20153DFC}, human-robot interaction~\citep{Dermy2017MultimodalIP}, and social robotics~\citep{MartnMartn2021JRDBAD}. In recent years, a variety of egocentric datasets and benchmarks have been established~\citep{Pirsiavash2012DetectingAO,Damen2020Collection,GarciaHernando2018FirstPersonHA,Tang2019MultiStreamDN,Dunnhofer2020IsFP}, and researchers have proposed various methods around the egocentric scene understanding, such as action recognition which summarizes an egocentric video clip into a certain action category~\citep{kazakos2019TBN,Sudhakaran2019LSTALS}, and action anticipation which infers future action types (without location information) based on the historical information~\citep{Liu2020ForecastingHI,Furnari2019WhatWY}. However, a fundamental egocentric vision problem remains underexplored: \textit{how to anticipate the future target location of someone's object manipulation action in 3D space}? This is crucial for more safe and effective planning and control of robots to collaborate with human. 

Basically, cognitive robots 
that need to interact with humans should be able to expect target locations of human actions at an early stage, allowing robots to compute appropriate reactions.
For example, when a person with upper-limb neuromuscular diseases is trying to grasp an object, a wearable exoskeleton should comprehend the human's intended target location before the grasping is completed, so the robot can plan its motion smoothly to help achieve the goal. Additionally, any computational latency in the prediction algorithm implementation could also be compensated if the robot is able to predict the action target ahead of time.

Thus, we establish a new dataset, \ourname, for egocentric prediction of action target in 3D. Our motivation lies in three aspects: \textbf{(1)} understanding the target locations of human actions is of significance to human-robot interaction~\citep{Kim2019EyesAF}, \textbf{(2)} prediction in 3D instead of 2D space facilitates the robot planning and control, and \textbf{(3)} there are no public datasets for egocentric action target prediction in 3D. To this end, we initiate the first study of \textit{egocentric 3D action target prediction}, which processes an egocentric sensor stream as an online signal and formulates action understanding as a continuous update for the target location. In summary, the main purpose of this study is to anticipate the 3D target locations of human actions as early as possible, to \textit{compensate for any latency} and \textit{support timely reactions} of robots.

Technically, it might seem too difficult if not impossible to ask the machine to accurately anticipate human intention locations especially at an early stage when there is very limited information. However human behaviors while attaining goal locations have certain distinct properties such as \textit{eyes are faster than hands}~\citep{Kim2019EyesAF}: when we try to grasp an object, we firstly search for the object in our visual field before reaching out for it. This phenomenon indicates that human intention locations could be anticipated based on the information of both \textit{visual perception} and \textit{head motion}. Therefore, our dataset  is multimodality, including RGB and depth images, and inertial measurement unit (IMU) data, which are all recorded by a helmet-mounted Azure Kinect RGB-D camera. In each recording, the camera wearer reaches for, grabs, and moves objects randomly placed in a household scene. Each recording features a different configuration of household objects within the scene. To annotate action targets less laboriously, we employ an off-the-shelf hand pose estimation model to localize the hand center which is used to denote the target location.

In order to solve this novel task of \textit{egocentric 3D action target prediction}, we propose a simple baseline approach on top of recurrent neural networks (RNNs) in conjunction with both visual and motion features. To summarize, our main contributions are as follows: 
\begin{itemize}[nosep,nolistsep]
    \item We initiate the first study of the egocentric action target prediction in 3D space.
    \item We build a novel \ourname~dataset and propose new evaluation metrics for this new egocentric vision task.
    \item We design a simple baseline method to achieve continuous prediction of action target, and comprehensively benchmark the performance.
    \item We open source all the code and dataset for reproducibility and future improvements.
\end{itemize}

\section{Related Work}\label{sec:related}

\begin{table*}[t]
\caption{\label{table:dataset} Comparison of datasets which can support research in \textbf{egocentric future prediction}. Prediction tasks are divided into three kinds: \textbf{(1)} action anticipation, \textbf{(2)} region prediction, and \textbf{(3)} trajectory/target prediction. The datasets listed in the \textbf{upper section} provide 2D cues, and the datasets in the \textbf{lower section} include 3D information such as depth and inertial measurement unit. N/P denotes Not Provided.}
\scriptsize
  \centering
   \setlength{\tabcolsep}{1.5mm}{
  \begin{tabular}{@{}c|ccccccccc@{}}
      \toprule
     \multirow{1}{*}{\bf Dataset} & {\bf Prediction Task} & {\bf Scenarios} &  {\bf Device} &  {\bf Modality} & {\bf Annotation} & {\bf \# of Frame} &  {\bf \# of Env.} &{\bf Year}  &{\bf Public}
      \\
    \midrule
    ADL~\citep{Pirsiavash2012DetectingAO} & Region & Manipulation & GoPro & RGB & Action/Object &
    1.0M & 20 & 2012 & \cmark \\
    GTEA Gaze$+$~\citep{Fathi2012LearningTR} & Region & Manipulation & SMI & RGB/Audio & Hand/Gaze &
    0.4M & 1 & 2012 & \cmark \\ 
    Daily Intentions Dataset~\citep{WuChienICCV17}& Action & Manipulation & Fisheye Len & RGB/IMU & Action & N/P & N/P & 2017 & \cmark
    \\
    EPIC-KITCHENS-50~\citep{Damen2018EPICKITCHENS} & Action/Region & Manipulation & GoPro & RGB/Audio & Action/Object &
    11.5M & 32 & 2018 & \cmark \\
    MAD~\citep{fermuller2018prediction}& Action & Manipulation & N/P & RGB/Force & Action &
    N/P & N/P & 2018 & \xmark \\

    ATT~\citep{Zhang2018FromCA} & Region & Manipulation & N/P & RGB & Gaze &
    217.0K & N/P & 2018 & \xmark \\
    EPIC-Tent~\citep{Jang2019EPICTentAE} & Region & Manipulation & GoPro/SMI & RGB/Audio & Gaze &
    1.2M & 1 & 2019 & \cmark \\
    100DoH~\citep{Shan20} & Region & Manipulation & YouTube & RGB & Hand &
    27.3K & N/P & 2020 & \cmark \\ 
    EPIC-KITCHENS-100~\citep{Damen2020Collection} & Action/Region & Manipulation & GoPro & RGB/Audio & Action/Object &
    20.0M & 45 & 2020 & \cmark \\
    EGTEA Gaze$+$~\citep{Li2021InTE} & Region & Manipulation & SMI & RGB/Audio & Hand/Gaze &
    2.4M & 1 & 2021 & \cmark \\  
    MECCANO~\citep{Ragusa2021TheMD} & Region & Manipulation & SR300 & RGB & Object & 0.3M & N/P & 2021 & \cmark
    \\
    Ego4D~\citep{grauman2021ego4d} & Action/Region & Manipulation & GoPro & RGB& Action/Object &
    N/P & N/P & 2021 & \cmark \\\midrule
    KrishnaCam~\citep{krishna-wacv2016} & Trajectory & Walking & Cellphone & IMU/GPS & Trajectory &
    7.6M & N/P & 2016 & \cmark \\  
    EgoMotion~\citep{Park2016EgocentricFL} & Trajectory & Walking & GoPro Stereo & RGB-D & Trajectory &
    65.5k & 26 & 2016 & \cmark \\  
    Ego4D~\citep{grauman2021ego4d} & Trajectory & Walking & Stereo & RGB-D & Trajectory &
    N/P & N/P & 2021 & \cmark \\  
    \textbf{\ourname~(ours)} & Target & Manipulation & Azure Kinect & RGB-D/IMU & Target &
    1M & 15 & 2021 & \cmark \\   
    \bottomrule
    \end{tabular}
    }
   \vspace{-6mm}
\end{table*}


\textbf{Egocentric datasets and benchmarks.}
Egocentric videos provide a wealth of knowledge on how humans see and interact with their surroundings, which is vital to understanding human behavior. The research in egocentric vision has been rapidly advanced owing to the development of wearable devices as well as egocentric datasets. In recent years, the scale of datasets has been gradually increased, and both scenes and annotations in egocentric scenarios have been enriched. For example, 2D object bounding boxes are provided to facilitate a variety of 2D computer vision tasks~\citep{Pirsiavash2012DetectingAO,Damen2014YouDoID,Damen2018EPICKITCHENS,Shan20,Dunnhofer2020IsFP}. Gaze measurements are supplemented to help understand the human intention in the image space~\citep{Fathi2012LearningTR,Zhang2018FromCA,Jang2019EPICTentAE,Li2021InTE}. Hand annotations are provided as a useful information to understand human-object interaction~\citep{Bambach2015LendingAH,GarciaHernando2018FirstPersonHA}. Thanks to these well-built datasets, various egocentric vision tasks have been proposed and studied such as action recognition~\citep{Li2021EgoExoTV, kazakos2019TBN,Sudhakaran2019LSTALS}, action anticipation~\citep{Liu2020ForecastingHI,Furnari2019WhatWY,girdhar2021anticipative}, video summarization~\citep{Lee2014PredictingIO, Lu2013StoryDrivenSF, Yonetani2016VisualMD}, hand-object interaction parsing~\citep{Cai2016UnderstandingHM, Nagarajan2019GroundedHI,Bandini2020AnalysisOT}, social interaction analysis~\citep{Fathi2012SocialIA, Ng2020You2MeIB, Yonetani2016RecognizingMA}, and egocentric object detection and tracking~\citep{Bambach2015LendingAH, Dunnhofer2020IsFP, Li2013ModelRW}. Despite numerous efforts to promote the development of egocentric vision, \textit{most datasets and tasks focus solely on 2D computer vision without 3D data}.
Ego4D~\citep{grauman2021ego4d}, a large-scale egocentric video dataset partially containing audio, mesh, stereo, and eye gaze information, was recently presented, and it also proposed five benchmarks centered on episodic memory, hands and objects, audio-visual diarization, social interactions, and activity forecasting. \textit{Despite its unprecedented scale and diversity, online 3D target location prediction in egocentric views remains insufficiently investigated}.

\textbf{Egocentric future prediction.}
In literature, there are substantial works in egocentric action recognition~\citep{Asnaoui2017ASO}, video summarization~\citep{Molino2017SummarizationOE}, hand analysis~\citep{Bandini2020AnalysisOT}, and future prediction~\citep{Rodin2021PredictingTF}. Here we only review the most relevant work, egocentric future prediction, which is a relatively new research area. Egocentric prediction of human activities, targets, and trajectories has wide applications such as assistive technologies~\citep{ohnbar2018personalized}, trajectory planning~\citep{Park2016EgocentricFL}, multimedia~\citep{Liang2015ARIH}, and robotics~\citep{Kim2019EyesAF}. There are mainly three streams in egocentric future prediction: \textbf{(1)} action anticipation, \textbf{(2)} region prediction, and \textbf{(3)} trajectory forecasting. The first two problems are intensively studied yet there are scarce works regarding the third one. Action anticipation aims to generate an action label given a historical video clip, and various datasets such as EPIC-KITCHENS~\citep{Damen2018EPICKITCHENS,Damen2020Collection} and EGTEA Gaze$+$~\citep{Li2021InTE} which can support action anticipation have promoted the research in this topic~\citep{Zhang2020AnEA,Furnari2020RollingUnrollingLF,Ke2019TimeConditionedAA,Gammulle2019PredictingTF}. Region prediction is to predict a 2D region on the image which will cover the human intended location in the future, and the target regions are denoted by object bounding boxes~\citep{Fan2018ForecastingHA,Bertasius2017FirstPA}, human-object interaction hotspots~\citep{Liu2020ForecastingHI}, or future gaze locations~\citep{Zhang2017DeepFG,Zhang2019AnticipatingWP}. Trajectory prediction attempts to forecast the future 
foot trajectories of humans~\citep{Bertasius2018EgocentricBM,Park2016EgocentricFL}. For example, Park \textit{et al.} proposed to generate plausible future trajectories of human ego-motion in egocentric stereo images~\citep{Park2016EgocentricFL}. Rhinehart \textit{et al.} used online inverse reinforcement learning to forecast a person’s walking destination and action in a 3D map~\citep{Rhinehart2020FirstPersonAF}. The recently-proposed Ego4D~\citep{grauman2021ego4d} developed a unified benchmark to evaluate the progress in egocentric future prediction including all the three prediction tasks (action/region/trajectory). However, the online prediction of action target in the 3D space still remains to be studied.

\textbf{Remark.} Existing datasets related to egocentric future prediction are summarized in Table~\ref{table:dataset}. In summary, most research in egocentric prediction have concentrated on the action category or 2D image region, and a few works have studied the egocentric 3D trajectory prediction in the walking scenarios. However, 3D action target prediction in the manipulation scenarios with rich hand-object interactions is still underexplored. Actually, target prediction is a special case of trajectory forecasting: the former only computes the end points of the trajectories. Yet in the manipulation scenarios, it is often infeasible to obtain complete hand trajectories because human hands often locate outside of the egocentric view. Therefore, we focus on predicting the 3D target location of an object manipulation action, which is desirable in robot planning and control.

\section{Egocentric Action Target Prediction in 3D}\label{sec:benchmark}

In this section, we define the problem of egocentric 3D action target prediction, discuss the challenges of the task, introduce the evaluation metrics for the proposed task, and present our simple baseline method.

\subsection{Problem Formulation}\label{subsec:problem}
\vspace{-1mm}
Many prior works in egocentric future prediction consider an offline mode, \textit{i.e.}, they require a fixed-length historical sequence to predict the future action or target region. Differently, we consider a more realistic online mode, \textit{i.e.}, our task requires online prediction based on variable-length historical information. Meanwhile, the difficulty level of our task varies in different temporal stages, \textit{e.g.}, in the early stage, the task is very challenging because \textbf{(1)} there is scarce information, and \textbf{(2)} the hands which can serve as crucial cues are often outside of the view. In contrast, the temporal information becomes rich and the hands are often clearly visible in the late stage, thus the task becomes easier over time. An example sequence is shown in Fig.~\ref{fig:task}.

\textbf{Notation.} The colored point cloud at frame $t$ denoted by ${\mathbf{X}_t} \in \mathbb{R}^{N_t \times 6}$ is represented as a set of 3D points $\{\mathbf{x}^n_t| n = 1,2,..., N_t\}$, where the $n$-th point of frame $t$ written as $\mathbf{x}^n_t \in \mathbb{R}^6$ is a vector of its Euclidean coordinate as well as RGB value $(x, y, z, r, g, b)$, and $N_t$ denotes the number of points at frame $t$. The IMU data at frame $t$ denoted by $\bm{\theta}_t \in \mathbb{R}^6$ is composed of the body-frame angular velocity $\bm{\omega}_t \in \mathbb{R}^3$ and linear acceleration $\bm{\alpha}_t \in \mathbb{R}^3$. The 3D target location at frame $t$ represented in the corresponding coordinate is denoted by $\bm{y}_t \in \mathbb{R}^3$.

For a clip with length $T$, the prediction will be executed  $T$ times to achieve continuous update for the action target in this clip. Note that the action target $\{\bm{y}_t | t=1,2,...,T \}$ within a clip is the same point in the world coordinate , yet has different values because they are represented in different local coordinates which depend on the head poses.

\textbf{Definition.}
Given historical colored point cloud streams $\mathbf{X}_{1:t} = \{\mathbf{X}_1, \mathbf{X}_2, ..., \mathbf{X}_t\}$ and IMU streams $\bm{\theta}_{1:t} = \{\bm{\theta}_1, \bm{\theta}_2, ..., \bm{\theta}_t\}$, we aim to predict the 3D target location $\bm{y}_t$ at each frame. From the machine learning perspective, $f$ denotes a model which takes visual  and motion sensor streams as input, and can output an estimation of future target locations:  ${\bm{o}}_t = f(\mathbf{X}_{1:t}, \bm{\theta}_{1:t})$. We seek an optimal $f$ to predict ${\bm{o}}_t$ as close to $\bm{y}_t $ as possible. Meanwhile, we prefer to generate accurate predictions as early as possible.

\textbf{Challenges.} There are three major challenges in \ourname. \textbf{(1)} \textit{Multi-modal information fusion}: the camera motion and visual information in a data stream have different importance for action target prediction at different action stages. Intuitively, the head motion appears to be more beneficial in the early action stage, while the visual appearance seems more useful in the late stage when the hand position start to be observed and can be exploited. It is nontrivial to effectively fuse these two types of information to achieve reliable prediction. \textbf{(2)} \textit{Early prediction}: achieving early stage prediction with high precision is usually more valuable for downstream applications such as robot control. But this is also difficult since information from the initial stage of an action is insufficient. \textbf{(3)} \textit{3D workspace}: predicting in 3D space further increases the task difficulty compared to predicting on a 2D image.

\textbf{Evaluation metrics.}
\textbf{(1)} \textit{Temporal-aware evaluation:} we divide each temporal window (an action clip) into ten stages, and calculate average center location error (CLE) for the predictions in each stage, so that we can observe the tendency of the prediction precision over time. \textbf{(2)} \textit{Early prediction evaluation:} we employ the prediction precision (CLE) when
observing only the beginning 10\%, 20\%, 30\%, 40\%, and 50\% of the action sequence, to assess the early prediction capability. \textbf{(3)} \textit{Overall evaluation:} we linearly weight the prediction errors based on the temporal stages: the prediction errors at the early stages are strongly penalized, and we can compute an overall score using temporal-aware weighted sum of the errors at different stages.
\vspace{-1mm}

\subsection{Baseline Method}
\begin{figure}[t]
\begin{center}
\includegraphics[width=0.46\textwidth]{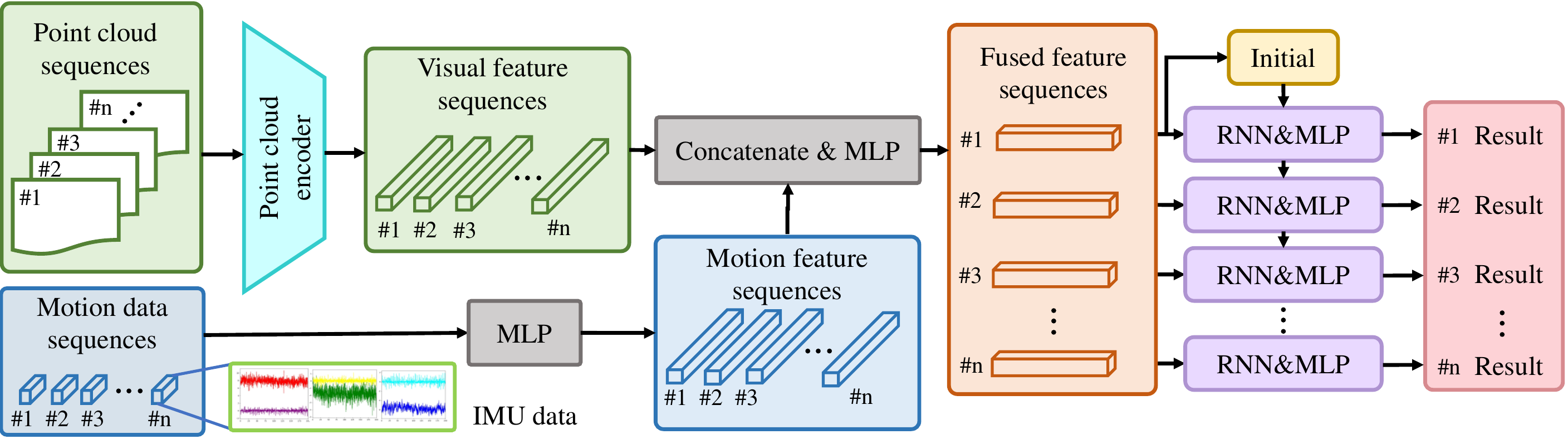}
\vspace{-5pt}
\caption{\textbf{Workflow of our baseline method.} The visual and motion features are separately extracted by two backbone networks, and then fused and fed into the RNN-based prediction module for the future action target localization.}
\label{fig:workflow}
\end{center} 
\vspace{-9mm}
\end{figure}

\vspace{-1mm}
As mentioned in Section~\ref{subsec:problem}, egocentric 3D action target prediction is a very challenging task. We propose a simple baseline method which uses two backbone networks separately for multimodality representation learning, followed by utilizing concatenation to achieve multimodality feature fusion, and we employ a recurrent neural network (RNN) to achieve continuous update for the 3D action target. The main workflow is presented in Fig.~\ref{fig:workflow}.

\textbf{Visual and motion features encoding.} We use a visual feature extractor denoted by $\psi$ which is based on a classic point cloud backbone PointConv~\citep{Wu2019PointConvDC}, to encode the visual features $\bm{v}_t   = \psi(\mathbf{X}_t)$. Besides, we use a motion feature extractor denoted by  $\phi$ which is based on multilayer perceptron (MLP) to encode the motion cues $\bm{m}_t = \phi(\bm{\theta}_t)$. After the feature encoding, the features of two modalities are concatenated and fed into another MLP to obtain the fused features $\bm{u}_t = \text{MLP}(\text{Cat}(\bm{v}_t, \bm{m}_t))$.

\textbf{Online 3D action target prediction.}
We divide the 3D space into grids and predict confidence value for each grid. We use a RNN to encode the sequential information and achieve online prediction, for example, at frame $t$, the score vector $\bm{s}_t \in \mathbb{R}^N$ for $N$-dimensional x-grids is computed by $\bm{s}_t = \text{RNN}(\bm{u}_t, \bm{h}_{t-1})$ (y and z directions are the same, so they are bypassed for simplicity), where  $\bm{h}_{t-1}$ is the learned hidden representation. RNN is able to learn both long- and short-term dependency in historical sensory streams which is desirable in our task.

\textbf{Training objective and loss function.} 
Since the adjoined grids are highly correlated compared to the general classification task, adopting the classic classification loss function (cross entropy loss) is not a promising choice, as demonstrated later in Section~\ref{alabtion}. Therefore, we redesign the loss function to utilize the dependencies between adjoined grids: we select the grid with a higher confidence score, and use the score to weight its importance. The comprehensive experiments prove that our loss, named as truncated weighted regression loss (TWRLoss), is more robust than the general one in this task which solely considers the single point with the highest confidence value. Mathematically, $\bm{g} \in \mathbb{R}^N $ denotes x-grids (we normalize the grid coordinate from -1 to 1), and the prediction score in the $n$-th x-grid is denoted by $\bm{s}_t[n] (n=0,1,...,N-1)$. Let $\bm{m}_l \in \mathbb{R}^N $ represent the binary mask for x-grids at frame $t$ to filter out grids with lower scores, and $\bm{m}_l[n]$ denote the binary value for the $n$-th x-grid, then the masked score in x-grids is calculated by:
\vspace{-5mm}
\begin{equation}
\begin{split}
    &\hat{\bm{s}}_t=\bm{m}_l \odot \bm{s}_t, \\
  \bm{m}_l[n]=
  &\left\{ 
    \begin{array}{ll}
    1,& n~\in~\{j~|~\bm{s}_t[j]>\gamma\}\\
    0, &  n~\in~\{j~|~\bm{s}_t[j]\leq \gamma\},
    \end{array}
    \right.\\
\end{split}
\end{equation}
where $\odot$ denotes the element-wise dot product, and $\gamma$ is a threshold to filter out the grids with low prediction scores. It is set to 0.5 in this method. Then the estimated target location's x-coordinate is calculated by: $p_t \in \mathbb{R} = \hat{\bm{s}}^T_t  \bm{g}$. Since the prediction difficulty is different as time goes on, the prediction error at different stages should be penalized differently, that is, we assign different weights to losses at different stages: $\mathcal{L} =  \sum_{t=1}^T w_t(y_t-p_t)^2$, where $y_t$ is the x-coordinate of the ground truth target point, and $w_t$ is a linear weight from 2 to 1, \textit{i.e.}, $w_t = 2 - \frac{t}{T}$.

\section{\ourname~Dataset}
\begin{figure*}[t]
\begin{center}
\hsize=\textwidth 
\includegraphics[width=0.98\textwidth]{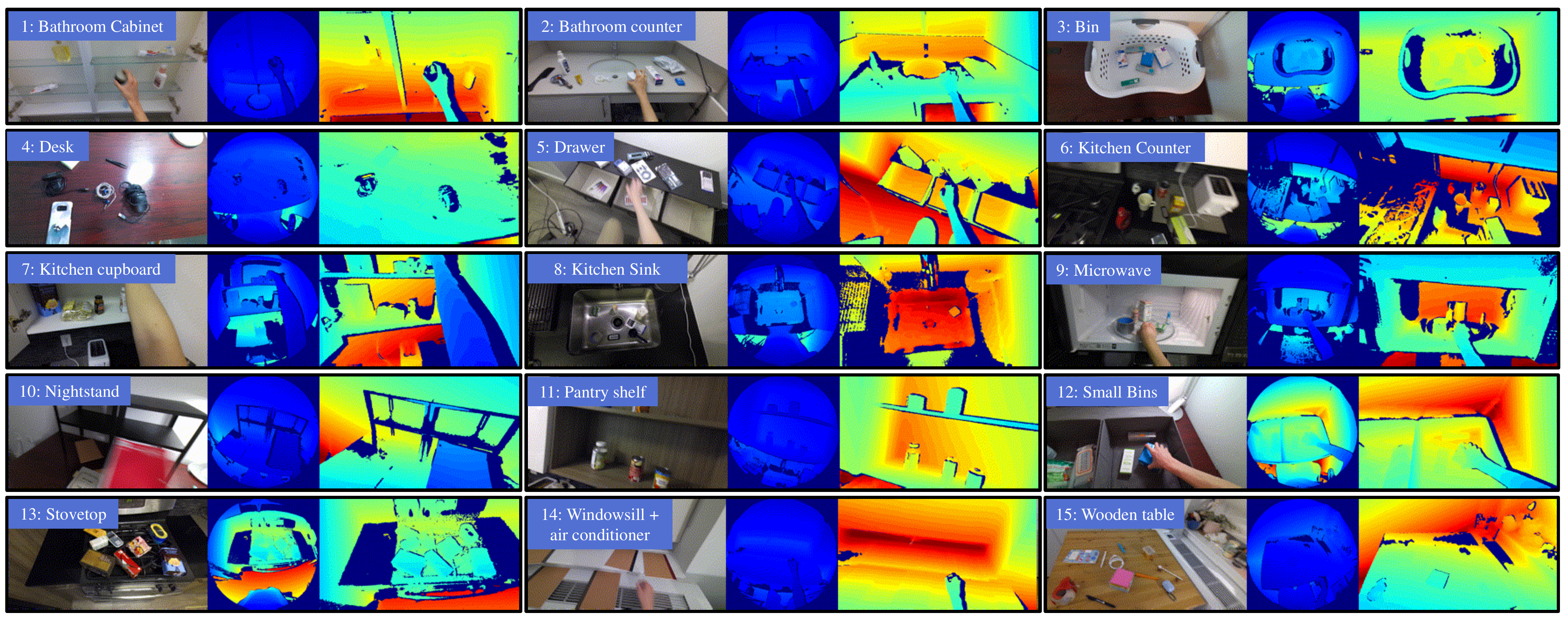}
\vspace{-2mm}
\caption{\textbf{Visualizations of scenes.} Left to right in three columns: RGB images, depth, depth transformed into RGB camera.}
\label{fig:scenes}
\end{center} 
\vspace{-9mm}
\end{figure*}

\subsection{Raw Data Acquisition}\label{subsec:raw}
\begin{figure}[t]
\begin{center}
\includegraphics[width=0.92\columnwidth]{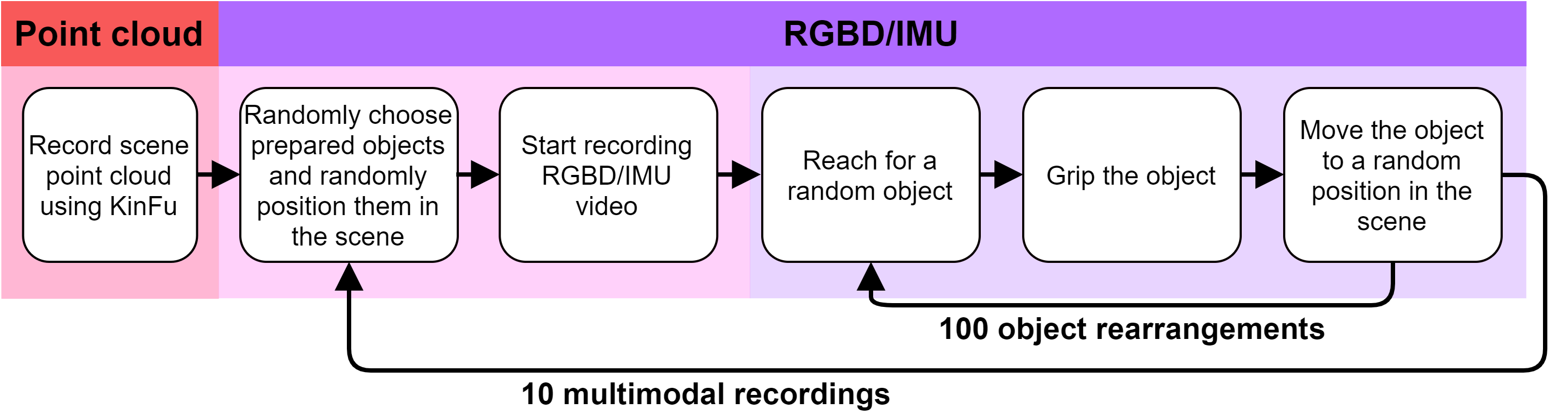}
\caption{\textbf{Illustration of the data collection pipeline.} In each scene, we randomly choose and place objects in the scene and then conduct object rearrangement for one hundred times.}
\label{fig:collection}
\end{center} 
\vspace{-10mm}
\end{figure}

The raw \ourname~data was recorded by 2 participants in 15 commonplace yet diverse household scenes. The recordings feature environments such as various kitchen, bedroom, and bathroom spaces. The participants continuously re-arranged the objects within the scene. Note that object rearrangement is a well-known task in the robotics community~\citep{Huang2019LargeScaleMR, Krontiris2015DealingWD, Levihn2012MultirobotMR}. We purposefully chose scenes in which hand-object manipulations, specifically grabbing and moving, often occur in everyday life (e.g. shelves, cabinets, counter tops, and other surfaces and fixtures where objects are stored or placed). All scenes are visualized in Fig.~\ref{fig:scenes}. The data collection procedure is demonstrated in Fig.~\ref{fig:collection}.

The raw data is captured in 15 scenes, and consists of 10 RGB-D/IMU recordings per scene. Meanwhile, a point cloud of the default state of 15 scenes (objects removed from the environment) is included. Each recording features approximately 100 hand-object actions and 4 minutes of footage at 30 frames per second (FPS) for both color and depth streams. The total collection contains 150 recordings, 15 household scene point clouds, 15,000 hand-object actions, 600 minutes of raw RGB-D/IMU data, 0.9 million hand-object action frames, and 1 million RGB-D frames for the entire dataset. All RGB-D/IMU data was collected using Microsoft's Azure Kinect DK and Azure Kinect Recorder software. The recordings were created with the following settings: 3840x2160 (4K) color camera resolution, 512x512 depth camera resolution, 30 fps for both color and depth streams, and Wide field-of-view depth mode (WFOV) 2x2 binned depth recording mode to more closely simulate the large field-of-view (FOV) of human vision. Depth delay was set to 0, and IMU recording was turned on. Scene point clouds were generated using OpenCV KinectFusion (KinFu) for Azure Kinect.

Data was acquired in sessions, during which a participant would be asked to record all the raw data corresponding to a selected scene in our dataset. No monetary compensation was offered to the participants; all recordings were performed on a purely voluntary basis, and no personally identifiable information was present in the process. Participants wore a helmet with an attached front-facing Azure Kinect camera, angled downward to capture an egocentric field-of-view and reliably record the participant's head and hand movements in all recordings during the session. The scene would be cleared before each session by removing nearby visible objects that could be manipulated by hand. A large collection of random items that can be grasped by hand, including but not limited to those that might realistically appear in the scene in daily situations (i.e. cooking utensils, dried foods, and spices for kitchen scenes), were prepared as objects that could later be placed in random configurations in the scene prior to each recording during the session. At the start of the session, the participant would be directed to follow a set of instructions:
\begin{enumerate}[nosep,nolistsep]
  \item Record a point cloud of the scene using KinFu installed on a nearby laptop.
  \item Adjust and wear helmet with mounted Azure Kinect so that the camera has a sufficient egocentric view of arm movements and hand-object interactions.
  \item Arbitrarily pick and arrange several of the prepared items in the scene.
  \item Stand or sit at designated location dependant on the scene, where any placed objects would easily be within field of view and arm's reach.
  \item Start RGB-D/IMU recording using Azure Kinect Recorder installed on a nearby laptop.
  \item Use a hand to re-arrange object in the scene 100 times.
  \item Stop recording and remove objects from the scene.
  \item Repeat steps 4-7 ten times for the scene, concluding the recording session.
  \vspace{-10pt}
\end{enumerate}

\subsection{Ground Truth Generation}\label{subsec:gt}
\vspace{-5pt}
Given a recording, we manually divide it into multiple action clips. To localize the 3D target in each clip, we use the following procedures. Firstly, we take the last frame of each clip based on the index provided by the manual division. Secondly, we use an off-the-shelf hand pose estimation model to localize the hand center in the last frame of each clip. Thirdly, we use colored point cloud registration to calculate the transformation matrices between the adjacent frames. Finally, for each clip, we transform the hand location in the last frame to historical frames according to the results of the third step, and the transformed locations can describe the 3D action target location in each frame's coordinate. Detailed procedures are presented as follows.

\begin{figure}[t]
\begin{center}
\includegraphics[width=0.47\textwidth]{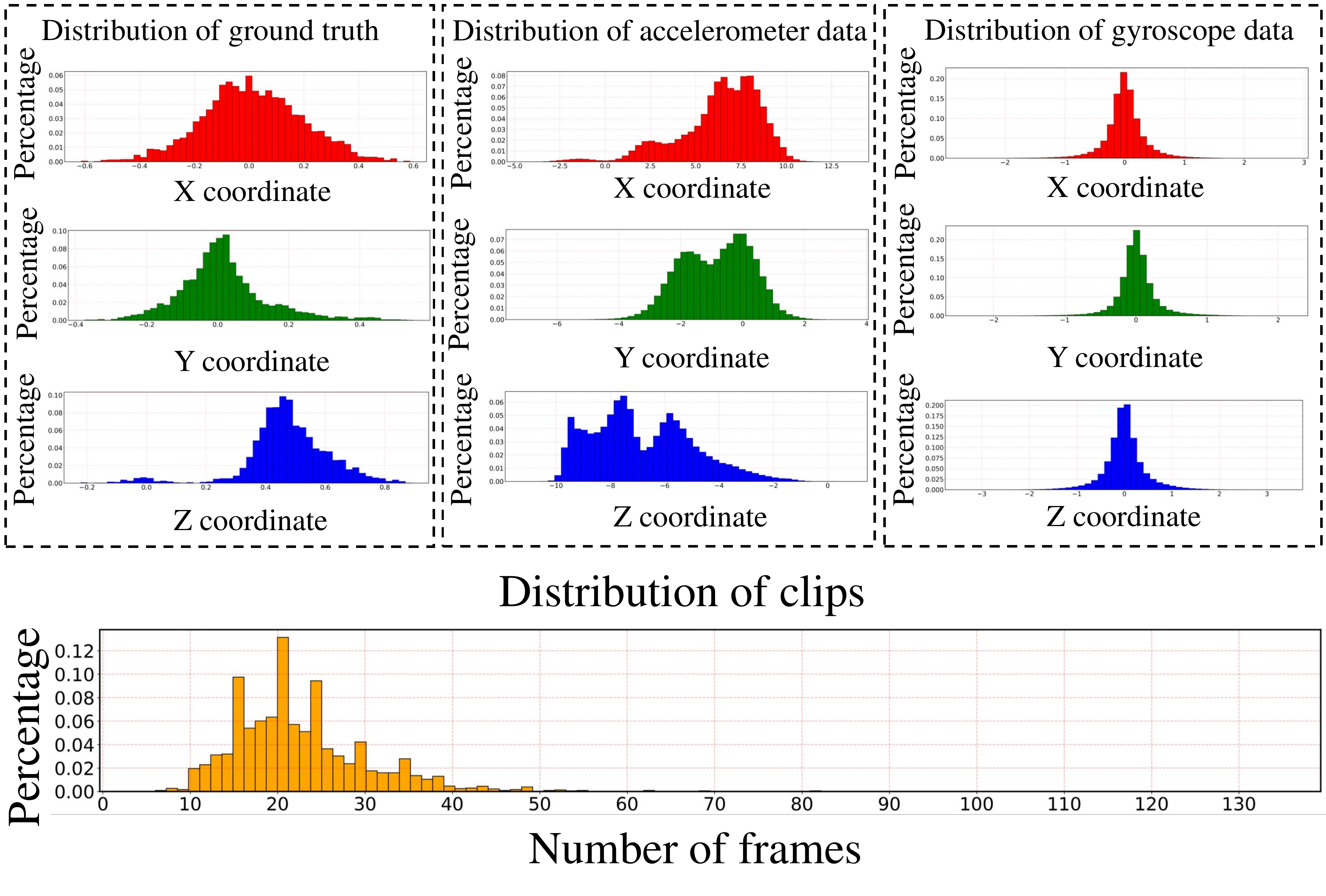}
\vspace{-10pt}
\caption{\textbf{Statistical properties of \ourname.} The distribution of 3D target locations and head motion (accelerometer and gyroscope) are visualized in xyz coordinate. The bottom figure shows the distribution of the number of frames in each action clip.}
\label{fig:dist}
\end{center} 
\vspace{-8mm}
\end{figure}

\textbf{Manual clip division.} We target short-term action target prediction, so we need to divide a long recording into multiple action clips such as reaching out for an object or placing an object. Specifically, we save the indexes of the first and the last frame for each clip, and such manual division is quite efficient: it takes around half an hour to manually annotate each recording. As shown in Fig.~\ref{fig:dist}, most action clips have 10-40 frames. After obtaining the index of the last frame for each clip, we use an off-the-shelf hand pose estimation model to localize the hand center, which is considered as the ground truth target.

\textbf{Hand pose estimation.}
For the last frame in each clip, 3D hand pose estimation is performed using Google's MediaPipe Hands python solution API, which first performs a single-shot palm detection task~\citep{Liu2016SSDSS} before  localization of 21 hand keypoints according to the MediaPipe hand landmark model. X and Y pixel coordinates of the keypoints were inferred this way, and the depth information (Z coordinate) of the hand is extracted from the corresponding depth frame transformed into color frame dimensions using the Azure Kinect SDK. Some hand pose estimation visualizations can be found in the supplementary.

\textbf{Visual odometry.} Since the camera (head) keeps moving when humans perform actions, the 3D target's coordinate is always changing although it is the same point in the world coordinate. We use colored Iterative Closest Point (ICP)~\citep{park2017colored} to compute the transformation matrices between two adjacent frames, which is usually called visual odometry. For each action clip, we can extract the hand location in the last frame to denote the target location for this action. Then we transform it into previous frames' coordinate system according to ICP. Therefore, the ground-truth action target in each frame could be generated. The distribution of the ground-truth target position is shown in Fig.~\ref{fig:dist}.

\section{Experiments}\label{sec:exp}
\subsection{Experimental Setup}
\begin{table*}[t]
	\scriptsize
	\centering
	\caption{\textbf{Quantitative results of different predictors on seen scenes.} Note that the lower score represents better performance. VF denotes visual features, TF denotes transformation matrices, IMU includes angular velocities as well as linear accelerations, r denotes the reverse of linear loss weight function (from linearly decreasing to linearly increasing). \textcolor[rgb]{ 1,  0,  0}{Red}, \textcolor[rgb]{ 0,  1,  0}{green}, and \textcolor[rgb]{ 0,  0,  1}{blue} fonts denote the top three performance.}
 	\vspace{-8pt}
 	\renewcommand\tabcolsep{7pt}
	\resizebox{0.82\linewidth}{!}{
		\begin{tabular}{l|c|ccccc|c cccc}
			\toprule
			\multirow{2}{*}{\textbf{Components}}& \multirow{2}{*}{\textbf{Overall (cm)}$\downarrow$} &\multicolumn{5}{c|}{\textbf{Early prediction (cm)}} &\multicolumn{5}{c}{\textbf{Late prediction (cm)}} \\
			 &   &10\%$\downarrow$ & 20\%$\downarrow$  & 30\%$\downarrow$ & 40\%$\downarrow$& 50\%$\downarrow$ &60\%$\downarrow$ & 70\%$\downarrow$  & 80\%$\downarrow$ & 90\%$\downarrow$  & 100\%$\downarrow$ \\
			\midrule
			
			   VF-r  & 19.21 & 23.67 & 22.02 & 20.58 & 19.27 & 18.16 & 17.29 & 16.69 & 16.50 & 16.42 & 16.63 \\
    VF    & 19.15 & \textcolor[rgb]{ 0,  0,  1}{\textbf{23.22}} & 21.80 & 20.98 & 19.79 & 18.32 & \textcolor[rgb]{ 0,  0,  1}{\textbf{17.29}} & \textcolor[rgb]{ 0,  0,  1}{\textbf{16.59}} & \textcolor[rgb]{ 0,  1,  0}{\textbf{16.08}} & \textcolor[rgb]{ 0,  0,  1}{\textbf{16.03}} & \textcolor[rgb]{ 0,  0,  1}{\textbf{16.24}} \\
    TF+IMU-r & \textcolor[rgb]{ 0,  1,  0}{\textbf{18.66}} & \textcolor[rgb]{ 1,  0,  0}{\textbf{22.84}} & \textcolor[rgb]{ 0,  1,  0}{\textbf{21.21}} & \textcolor[rgb]{ 1,  0,  0}{\textbf{19.94}} & \textcolor[rgb]{ 0,  1,  0}{\textbf{18.76}} & \textcolor[rgb]{ 0,  1,  0}{\textbf{17.68}} & \textcolor[rgb]{ 0,  1,  0}{\textbf{16.93}} & \textcolor[rgb]{ 0,  1,  0}{\textbf{16.33}} & \textcolor[rgb]{ 0,  0,  1}{\textbf{16.08}} & \textcolor[rgb]{ 0,  1,  0}{\textbf{15.99}} & 16.24 \\
    TF+IMU & 19.81 & 23.97 & 22.16 & 20.90 & 19.76 & 18.79 & 18.09 & 17.69 & 17.55 & 17.45 & 17.54 \\
    VF+IMU-r & 20.87 & 24.70 & 23.40 & 22.37 & 21.08 & 19.83 & 19.12 & 18.67 & 18.39 & 18.31 & 18.41 \\
    VF+IMU & 21.81 & 25.36 & 24.33 & 23.28 & 22.39 & 21.34 & 20.36 & 19.49 & 19.06 & 18.85 & 18.90 \\
    VF+TF-r & 20.41 & 23.50 & 23.51 & 22.81 & 21.53 & 20.06 & 18.86 & 17.77 & 17.09 & 16.74 & 16.69 \\
    VF+TF & 19.48 & 23.47 & 22.31 & 21.25 & 20.09 & 18.87 & 17.92 & 16.95 & 16.32 & 16.11 & \textcolor[rgb]{ 0,  1,  0}{\textbf{16.18}} \\
    VF+TF+IMU-r & \textcolor[rgb]{ 0,  0,  1}{\textbf{18.97}} & \textcolor[rgb]{ 0,  1,  0}{\textbf{23.02}} & \textcolor[rgb]{ 1,  0,  0}{\textbf{21.11}} & \textcolor[rgb]{ 0,  1,  0}{\textbf{20.01}} & \textcolor[rgb]{ 0,  0,  1}{\textbf{19.03}} & \textcolor[rgb]{ 0,  0,  1}{\textbf{18.13}} & 17.40 & 16.85 & 16.65 & 16.58 & 16.81 \\
    VF+TF+IMU & \textcolor[rgb]{ 1,  0,  0}{\textbf{18.61}} & 23.73 & \textcolor[rgb]{ 0,  0,  1}{\textbf{21.78}} & \textcolor[rgb]{ 0,  0,  1}{\textbf{20.20}} & \textcolor[rgb]{ 1,  0,  0}{\textbf{18.65}} & \textcolor[rgb]{ 1,  0,  0}{\textbf{17.37}} & \textcolor[rgb]{ 1,  0,  0}{\textbf{16.43}} & \textcolor[rgb]{ 1,  0,  0}{\textbf{15.77}} & \textcolor[rgb]{ 1,  0,  0}{\textbf{15.47}} & \textcolor[rgb]{ 1,  0,  0}{\textbf{15.43}} & \textcolor[rgb]{ 1,  0,  0}{\textbf{15.67}} \\
     \midrule
    NLLLoss-r & 26.12 & 29.69 & 28.04 & 27.30 & 26.30 & 25.37 & 24.83 & 24.27 & 23.94 & 23.77 & 23.76 \\
    NLLLoss & \textcolor[rgb]{ 0,  0,  1}{\textbf{21.63}} & \textcolor[rgb]{ 0,  0,  1}{\textbf{26.45}} & \textcolor[rgb]{ 0,  0,  1}{\textbf{23.90}} & \textcolor[rgb]{ 0,  0,  1}{\textbf{22.46}} & \textcolor[rgb]{ 0,  0,  1}{\textbf{21.36}} & \textcolor[rgb]{ 0,  0,  1}{\textbf{20.44}} & \textcolor[rgb]{ 0,  0,  1}{\textbf{19.88}} & \textcolor[rgb]{ 0,  0,  1}{\textbf{19.46}} & \textcolor[rgb]{ 0,  0,  1}{\textbf{19.30}} & \textcolor[rgb]{ 0,  0,  1}{\textbf{19.25}} & \textcolor[rgb]{ 0,  0,  1}{\textbf{19.41}} \\
    TWRLoss-r & \textcolor[rgb]{ 0,  1,  0}{\textbf{18.97}} & \textcolor[rgb]{ 1,  0,  0}{\textbf{23.02}} & \textcolor[rgb]{ 1,  0,  0}{\textbf{21.11}} & \textcolor[rgb]{ 1,  0,  0}{\textbf{20.01}} & \textcolor[rgb]{ 0,  1,  0}{\textbf{19.03}} & \textcolor[rgb]{ 0,  1,  0}{\textbf{18.13}} & \textcolor[rgb]{ 0,  1,  0}{\textbf{17.40}} & \textcolor[rgb]{ 0,  1,  0}{\textbf{16.85}} & \textcolor[rgb]{ 0,  1,  0}{\textbf{16.65}} & \textcolor[rgb]{ 0,  1,  0}{\textbf{16.58}} & \textcolor[rgb]{ 0,  1,  0}{\textbf{16.81}} \\
    TWRLoss & \textcolor[rgb]{ 1,  0,  0}{\textbf{18.61}} & \textcolor[rgb]{ 0,  1,  0}{\textbf{23.73}} & \textcolor[rgb]{ 0,  1,  0}{\textbf{21.78}} & \textcolor[rgb]{ 0,  1,  0}{\textbf{20.20}} & \textcolor[rgb]{ 1,  0,  0}{\textbf{18.65}} & \textcolor[rgb]{ 1,  0,  0}{\textbf{17.37}} & \textcolor[rgb]{ 1,  0,  0}{\textbf{16.43}} & \textcolor[rgb]{ 1,  0,  0}{\textbf{15.77}} & \textcolor[rgb]{ 1,  0,  0}{\textbf{15.47}} & \textcolor[rgb]{ 1,  0,  0}{\textbf{15.43}} & \textcolor[rgb]{ 1,  0,  0}{\textbf{15.67}} \\
    \midrule
    
    GRU-based-r & \textcolor[rgb]{ 0,  1,  0}{\textbf{18.64}} & \textcolor[rgb]{ 1,  0,  0}{\textbf{22.79}} & \textcolor[rgb]{ 0,  1,  0}{\textbf{21.29}} & \textcolor[rgb]{ 0,  1,  0}{\textbf{20.05}} & \textcolor[rgb]{ 0,  1,  0}{\textbf{18.82}} & \textcolor[rgb]{ 0,  1,  0}{\textbf{17.64}} & \textcolor[rgb]{ 0,  1,  0}{\textbf{16.79}} & \textcolor[rgb]{ 0,  1,  0}{\textbf{16.25}} & \textcolor[rgb]{ 0,  1,  0}{\textbf{16.02}} & \textcolor[rgb]{ 0,  1,  0}{\textbf{15.90}} & \textcolor[rgb]{ 0,  1,  0}{\textbf{16.16}} \\
    GRU-based & 20.07 & 23.73 & 22.10 & 21.03 & 20.04 & 19.10 & 18.59 & 18.19 & 17.97 & 18.01 & 18.27 \\
    LSTM-based-r & \textcolor[rgb]{ 0,  0,  1}{\textbf{18.97}} & \textcolor[rgb]{ 0,  1,  0}{\textbf{23.02}} & \textcolor[rgb]{ 1,  0,  0}{\textbf{21.11}} & \textcolor[rgb]{ 1,  0,  0}{\textbf{20.01}} & \textcolor[rgb]{ 0,  0,  1}{\textbf{19.03}} & \textcolor[rgb]{ 0,  0,  1}{\textbf{18.13}} & \textcolor[rgb]{ 0,  0,  1}{\textbf{17.40}} & \textcolor[rgb]{ 0,  0,  1}{\textbf{16.85}} & \textcolor[rgb]{ 0,  0,  1}{\textbf{16.65}} & \textcolor[rgb]{ 0,  0,  1}{\textbf{16.58}} & \textcolor[rgb]{ 0,  0,  1}{\textbf{16.81}} \\
    LSTM-based & \textcolor[rgb]{ 1,  0,  0}{\textbf{18.61}} & \textcolor[rgb]{ 0,  0,  1}{\textbf{23.73}} & \textcolor[rgb]{ 0,  0,  1}{\textbf{21.78}} & \textcolor[rgb]{ 0,  0,  1}{\textbf{20.20}} & \textcolor[rgb]{ 1,  0,  0}{\textbf{18.65}} & \textcolor[rgb]{ 1,  0,  0}{\textbf{17.37}} & \textcolor[rgb]{ 1,  0,  0}{\textbf{16.43}} & \textcolor[rgb]{ 1,  0,  0}{\textbf{15.77}} & \textcolor[rgb]{ 1,  0,  0}{\textbf{15.47}} & \textcolor[rgb]{ 1,  0,  0}{\textbf{15.43}} & \textcolor[rgb]{ 1,  0,  0}{\textbf{15.67}} \\
    
			\bottomrule
	\end{tabular}}
	\label{tab:seen}%
 	\vspace{-15pt}
\end{table*}%

\begin{table}[t]
	\scriptsize
	\centering
	\caption{\textbf{Quantitative comparison of different predictors on unseen scenes.} The meanings of VF, TF, and IMU are the same as Table~\ref{tab:seen}.}
 	\vspace{-8pt}
 	\renewcommand\tabcolsep{0.5pt}
	\resizebox{1\linewidth}{!}{
		\begin{tabular}{l|c|ccccc|c cccc}
			\toprule
			\multirow{2}{*}{\textbf{Comp.}}& \multirow{2}{*}{\textbf{Over. (cm)}$\downarrow$} &\multicolumn{5}{c|}{\textbf{Early prediction (cm)}} &\multicolumn{5}{c}{\textbf{Late prediction (cm)}} \\
			 &   &10\%$\downarrow$ & 20\%$\downarrow$  & 30\%$\downarrow$ & 40\%$\downarrow$& 50\%$\downarrow$ &60\%$\downarrow$ & 70\%$\downarrow$  & 80\%$\downarrow$ & 90\%$\downarrow$  & 100\%$\downarrow$ \\
			\midrule
			
			   VF    & 19.61 & 24.19 & 22.56 & 20.76 & 19.22 & 18.34 & 17.80 & 17.31 & 17.02 & 16.98 & 17.25 \\
    TF+IMU & 19.92 & 23.37 & 21.94 & 20.85 & 19.86 & 19.11 & 18.61 & 18.16 & 17.87 & 17.80 & 17.97 \\
    VF+IMU & 19.54 & 23.67 & 22.07 & 20.83 & 19.75 & 18.70 & 17.98 & 17.37 & \textbf{16.96} & \textbf{16.68} & \textbf{16.72} \\
    VF+TF & 20.82 & 24.38 & 23.22 & 21.93 & 20.82 & 19.92 & 19.28 & 18.85 & 18.62 & 18.59 & 18.68 \\
    VF+TF+IMU & \textbf{18.82} & \textbf{22.75} & \textbf{20.38} & \textbf{19.26} & \textbf{18.55} & \textbf{18.07} & \textbf{17.64} & \textbf{17.25} & {16.97} & {16.92} & {17.04} \\
    \midrule
    NLLLoss & 21.71 & 25.74 & 23.63 & 22.46 & 21.58 & 20.89 & 20.40 & 19.90 & 19.60 & 19.48 & 19.62 \\
    TWRLoss & \textbf{18.82} & \textbf{22.75} & \textbf{20.38} & \textbf{19.26} & \textbf{18.55} & \textbf{18.07} & \textbf{17.64} & \textbf{17.25} & \textbf{16.97} & \textbf{16.92} & \textbf{17.04} \\
    \midrule
    GRU-based & 19.77 & 23.32 & 21.81 & 20.77 & 19.75 & 18.96 & 18.44 & 17.94 & 17.67 & 17.55 & 17.74 \\
    LSTM-based & \textbf{18.82} & \textbf{22.75} & \textbf{20.38} & \textbf{19.26} & \textbf{18.55} & \textbf{18.07} & \textbf{17.64} & \textbf{17.25} & \textbf{16.97} & \textbf{16.92} & \textbf{17.04} \\
			\bottomrule
	\end{tabular}}
	\label{tab:unseen}%
 	\vspace{-20pt}
\end{table}%

\textbf{Dataset preparation.} We use 11 scenes in our experiments, \textit{i.e.}, 5 seen scenes: bathroom cabinet, bathroom counter, drawer, kitchenCounter, nightstand, and 6 unseen scenes: bin, kitchen cupboard, microwave, stovetop, windowsill+air conditioner, and wooden table. Our training, validation, and test sets are composed of 1990, 358, 314 action clips respectively, and the unseen test set which is employed to test generalization ability contains 772 action clips from 10 scenes. The length of each action clip ranges from 6 frames to 133 frames (23 frames on average).

\textbf{Implementation details}. During the training of 30 epoches, the SGD optimizer is employed with an initial learning rate of 0.01 and a decay factor of 0.9 for every 5 epochs. Besides, the momentum, weight decay of SGD, and batch size are set to 0.9, $10^{-4}$, and 8 respectively. In our baseline method, we adopt two RNN layers (LSTM or GRU) to exploit the temporal information. Our action target predictor is trained on NVIDIA GeForce RTX 3090 GPUs. We report the results on the test set.

\subsection{Quantitative Results}~\label{alabtion}
The temporal-aware quantitative results  are presented in Table~\ref{tab:seen}. In the seen scenes, the error can be decreased from  $\sim$ $25$cm to $\sim$ $15$cm, which is around the average length of an adult male's hand. Therefore, our baseline approach achieves a satisfactory performance in the task of egocentric prediction for action target. Meanwhile, there is still a remarkable gap to fill, so the proposed task is worthy of further investigation. The experimental results on unseen scenes validate the generalization ability of our baseline method, as shown in Table~\ref{tab:unseen}.

\textbf{Discussions on features}. To test the effect of different features on the performance, ablation studies on input features are performed, and the quantitative results are shown in Table~\ref{tab:seen}. We see that: \textbf{(1)} the method equipped with all the features achieves the best performance, validating the significance of multimodality features; \textbf{(2)} combining visual features, translation, rotation, velocity and acceleration together (VF+TF+IMU) can achieve excellent performance, while missing first-order (VF+IMU) or higher-order (VF+TF) motion features leads to worse performance than the method using visual features only, proving that the first-order and higher-order motion features are both crucial.

\textbf{Discussions on loss weight}. We also compared the performance of the models trained with two different linear loss weight functions: \textbf{(1)} linearly reduced from 2 to 1 (stronger penalty at the start), and \textbf{(2)} linearly grew from 1 to 2 (stronger penalty at the end). We find that implementing a harsher penalty early on could improve performance.

\textbf{Discussions on loss function}. Compared to Negative Log Likelihood Loss (NLLLoss), our truncated weighted regression loss (TWRLoss) incorporate the grids with higher scores into our training objective, thereby achieving better overall performance in both seen scenes (21.63$\xrightarrow{}$18.61) and unseen scenes (21.71$\xrightarrow{}$18.82). Meanwhile, the late prediction capability is also improved, for example, in seen scenes, the error could be decreased from 19.41 to 15.67 when observing 100\% data. 

\textbf{Discussions on RNN}. LSTM can promote the overall performance a little bit compared to GRU (20.07$\xrightarrow{}$18.61 in seen scenes and 19.77$\xrightarrow{}$18.82 in novel scenes). This is maybe because LSTM has more learnable parameters than GRU: GRU's bag has two gates (reset and update) while LSTM has three (input, output, forget). A more appropriate architecture design may further improve the prediction.
\begin{figure*}[t]
\centering
\includegraphics[width=0.95\textwidth]{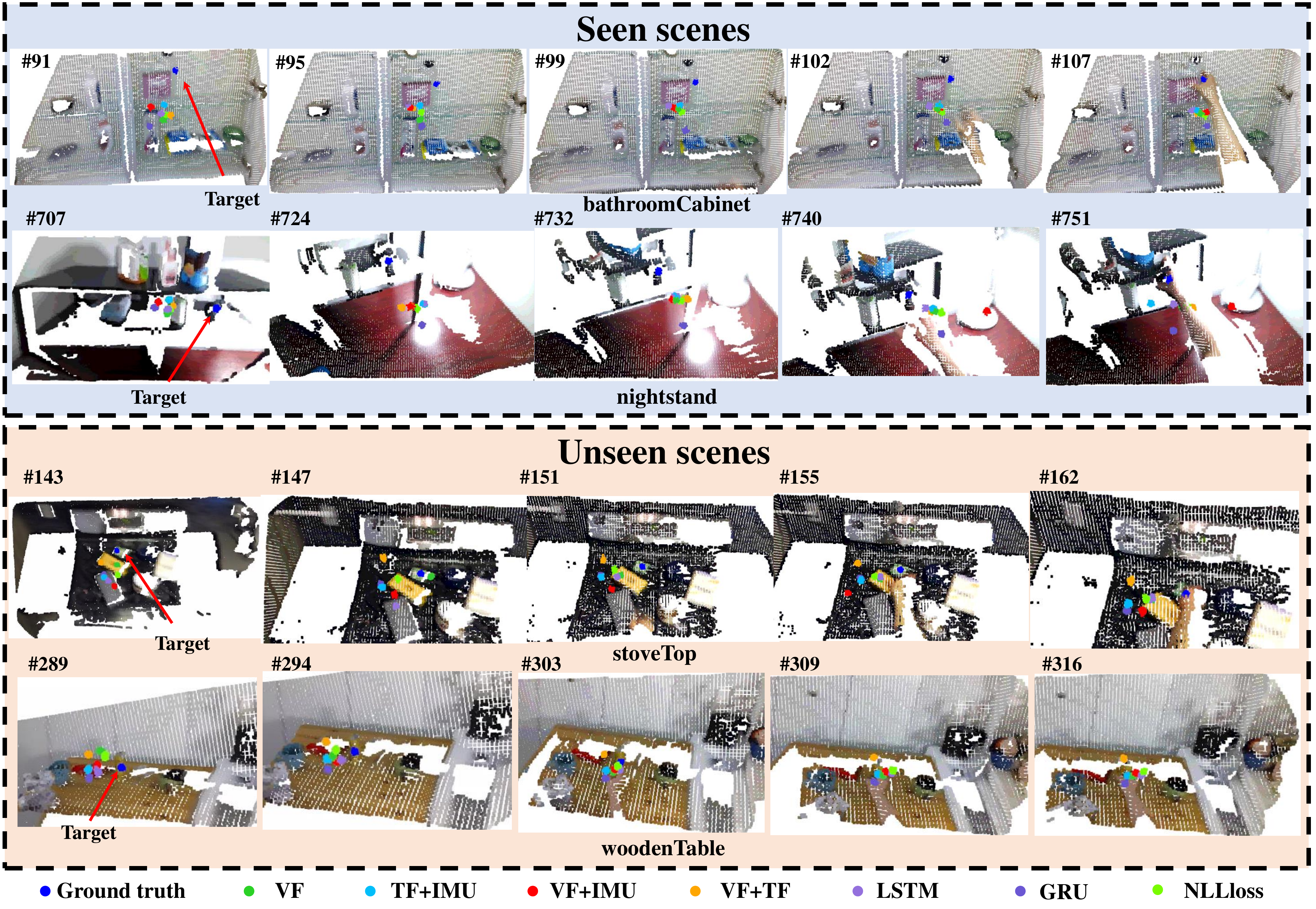}
\vspace{-2mm}
\caption{\textbf{Qualitative evaluation of different predictors in seen and unseen scenes.} The \textcolor[rgb]{0,0,1}{blue} point represents the ground-truth target.}
\label{fig:quali_seen}
\vspace{-5mm}
\end{figure*}

\textbf{Ablation study on the grid granularity. } The performance gradually improves when the granularity is increased, as shown in Table~\ref{tab:ablation}.

\begin{table}[t]
	\scriptsize
	\centering
	\caption{Ablation studies on the granularity of the grid. }
	 	\vspace{-8pt}
 	\renewcommand\tabcolsep{1pt}
	\resizebox{1\linewidth}{!}{
		\begin{tabular}{l|c|ccccc|c cccc}
			\toprule
			\multirow{2}{*}{\textbf{Granularity}}& \multirow{2}{*}{\textbf{Over.}$\downarrow$} &\multicolumn{5}{c|}{\textbf{Early prediction}} &\multicolumn{5}{c}{\textbf{Late prediction}} \\
			 &   &10\%$\downarrow$ & 20\%$\downarrow$  & 30\%$\downarrow$ & 40\%$\downarrow$& 50\%$\downarrow$ &60\%$\downarrow$ & 70\%$\downarrow$  & 80\%$\downarrow$ & 90\%$\downarrow$  & 100\%$\downarrow$ \\
			\midrule
			
			$1024^3/m^3$ & 19.88 & 24.03 & 22.08 & 20.69 & 19.70 & 18.86 & 18.30 & 17.91 & 17.71 & 17.66 & 17.86 \\
    $3072^3/m^3$ & 19.04 & 23.36 & 21.46 & 20.23 & 19.15 & 18.14 & 17.31 & 16.64 & 16.41 & 16.43 & 16.73 \\
    $5120^3/m^3$ & 18.61 & 23.73 & 21.78 & 20.20 & 18.65 & 17.37 & 16.43 & 15.77 & 15.47 & 15.43 & 15.67 \\

			\bottomrule
	\end{tabular}}
	\label{tab:ablation}%
 	\vspace{-20pt}
\end{table}%

\subsection{Qualitative Results}
\vspace{-1mm}
We visualize some prediction results in seen scenes and unseen scenes respectively in Fig.~\ref{fig:quali_seen}. We can find that when more temporal information is accumulated, the predictors can generate more precise results. Meanwhile, our baseline approaches demonstrate satisfactory generalization ability. However, there is still a notable gap to be discussed next.

\subsection{Limitations and Future works}\label{subsec:limitation}
\textbf{Dataset.}
As the initial phase of this academic research, due to limited resources and time, we do not have diverse demographics of skin color, age, weight, height, etc., for our dataset collection participants, nor comprehensive action types. This could limit the generalization ability of models learned on this dataset for any real-world products. However, we believe \textit{the significance of this research is to initiate a useful dataset for relevant academic communities to more easily start to work together on this novel task that will eventually improve life quality of people with disabilities}. Therefore the \textit{initial version of the dataset should not need to be ready for industry usage}. Moreover, as a common practice in datasets of this community, \textit{this paper is not the end of this study} and we are committed to continue growing the dataset to improve the diversity.

\textbf{Baseline.}
Our simple RNN baseline achieves a reasonable performance, although not accurate enough: a 20~cm prediction error could still lead to an unintended grasp of a wrong object on the table for a wearable robot user. This could be due to that the multimodality data is only fused with a naive concatenation, which may not be enough to distinguish the importance of different modality at different timestamps. To overcome this limitation, temporal-aware multimodality learning could be a future direction. Moreover, the exploitation of temporal information can also be improved: transformer structure might be more effective than RNN to deal with long sequences. We hope the relevant communities could address these limitations together and improve this task further so as to enable better human-robot collaboration.

\textbf{Potential negative social impacts.} 
Although our intention of proposing this new task and dataset is to improve human-machine interaction by predicting human intentions, which could help people with disabilities, it is not difficult for cyberpunk Sci-Fi writers to plot evil usages of this to-be-developed technology for building robotic soldiers that are unbeatable by humans. In addition, developing deep learning models has been criticized for its high power consumption and negative impact on climate change.

\vspace{-1mm}
\section{Conclusion}\label{sec:conclusion}
\vspace{-1mm}
In this work, we propose the first 3D dataset and benchmark for egocentric prediction of action target, which could play a crucial role in wearable devices, human-robot interaction, and augmented reality. Our annotation can be semi-automatic with several off-the-shelf machine learning algorithms, thus is quite efficient. Meanwhile,  we design a simple baseline approach based on RNNs to solve the novel task, which is the first method to localize the future action target in 3D. We believe our dataset and benchmark are useful for vision, robotics, and learning communities.

\textbf{Acknowledgement.}
The research is supported by NSF FW-HTF program under DUE-2026479. The authors gratefully acknowledge the constructive comments and suggestions from the anonymous reviewers.

{\small
\bibliographystyle{ieee_fullname}
\bibliography{egbib}
}

\renewcommand{\thetable}{\Roman{table}}
\renewcommand{\thefigure}{\Roman{figure}}
\renewcommand\thesection{\Roman {section}}

\section*{Appendix}
\setcounter{section}{0}
\setcounter{figure}{0}
\setcounter{table}{0}




\section{Data collection}
The helmet we chose to mount the Azure Kinect RGB-D camera to is a bicycle helmet suitable for head circumferences between 52 and 56 cm, as shown in Fig.~\ref{fig:datacollection} in the supplementary. The helmet itself has adjustable straps and an inner foam shell for comfort. No parts of the helmet appeared in the wearer’s field of view.

We limited data collection to 100 actions per recording (approximately 4 minutes to complete) to reduce the possibility of fatigue from wearing the helmet (405 grams) with the Azure Kinect RGB-D camera (440 grams).

Additionally, all cables used during recordings were long enough so that they would not restrict the data collector’s movements. The cables were also positioned in a way that would not reduce visibility during recordings: The USB cable for supplying power to the camera was plugged into a 15-feet long extension cord behind the data collector, and the laptop used for recording data was placed as close as possible to the data collector while not obstructing the data collector’s view of the rest of the scene. The USB-C data cable connecting the camera to the laptop was also positioned behind the data collector for the same reason as the power cable.

Thanks to the above setup, it does not seem that the volunteers' movement patterns have any noticeable change according to our observation.

\begin{figure}[b]
    \centering
    \includegraphics[width=0.48\textwidth]{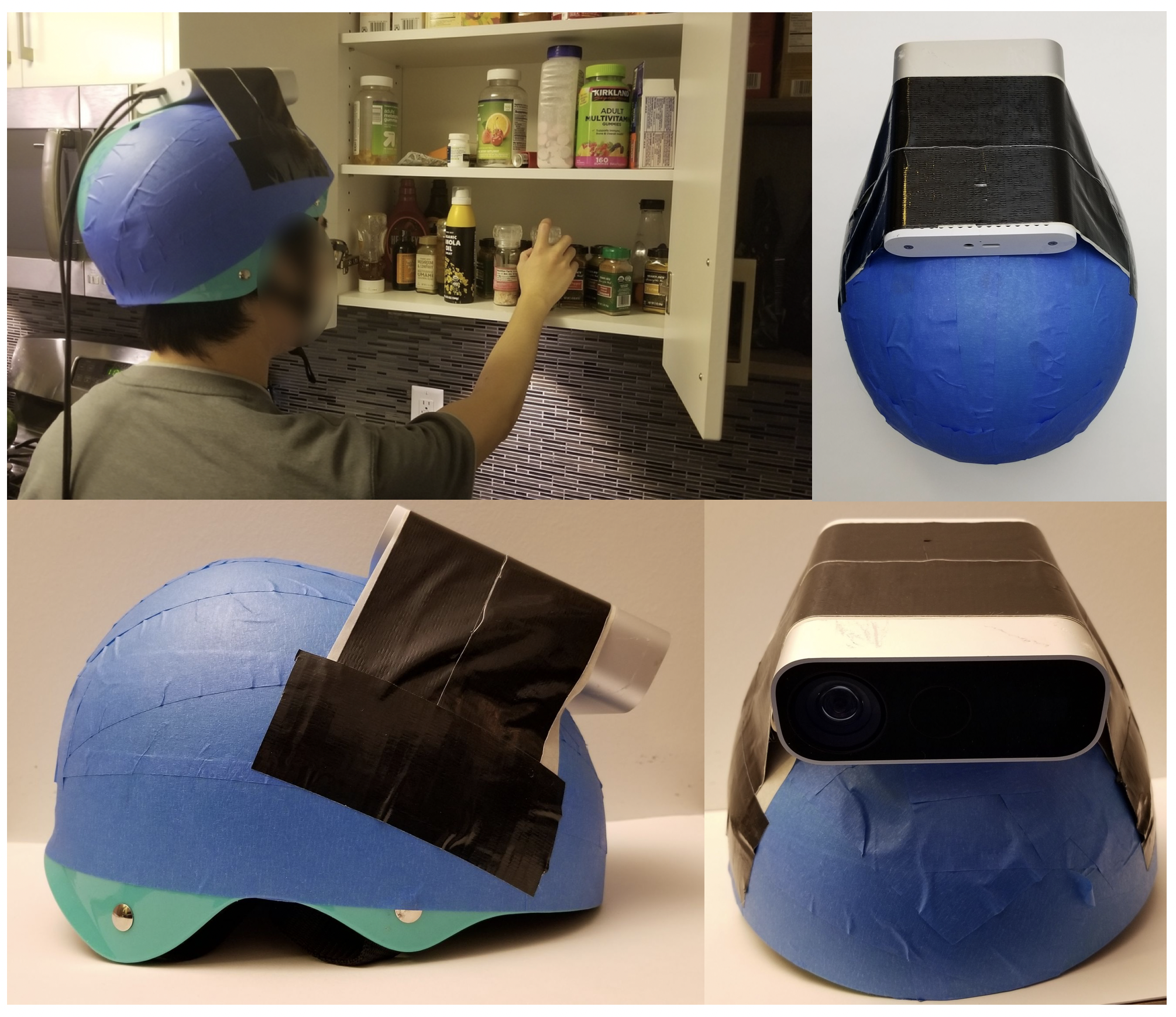}
    \caption{{Photos of helmets.}}
    \label{fig:datacollection}
\end{figure}

\section{Visualizations of hand pose estimation}
\begin{figure}[t]
    \centering
    \includegraphics[width=0.48\textwidth]{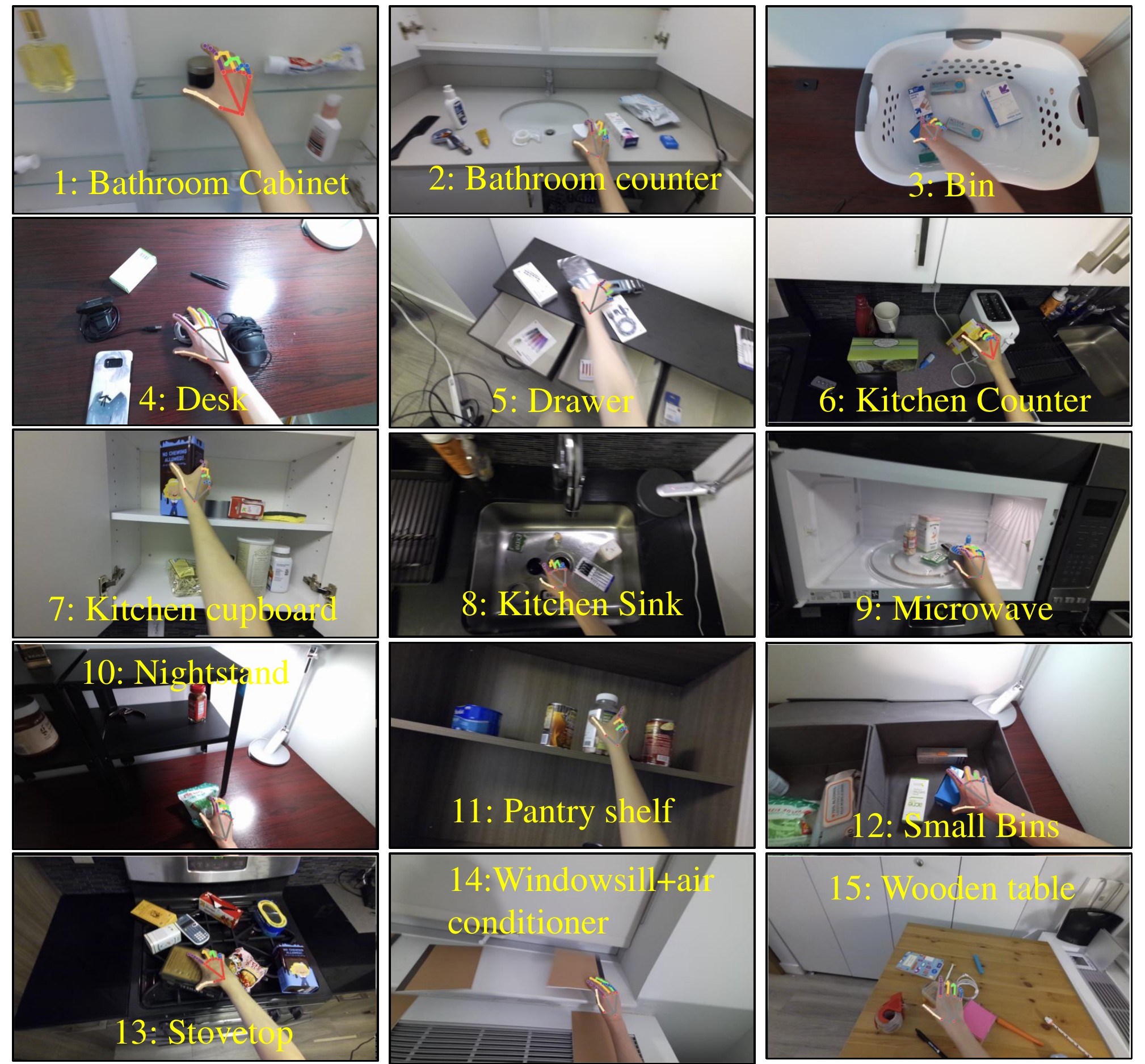}
    \caption{{Visualizations of hand pose estimations.}}
    \label{fig:handpose}
\end{figure}

We conducted a quality check for hand pose estimation results, and only the frames with reliable estimations were included in our dataset. Google's MediaPipe Hands tool was quite robust, as most frames had received precise estimations. Some example frames are presented in Fig.~\ref{fig:handpose}.

\section{Failure cases} 
\begin{figure}[t]
  \centering
  \includegraphics[width=1.0\linewidth]{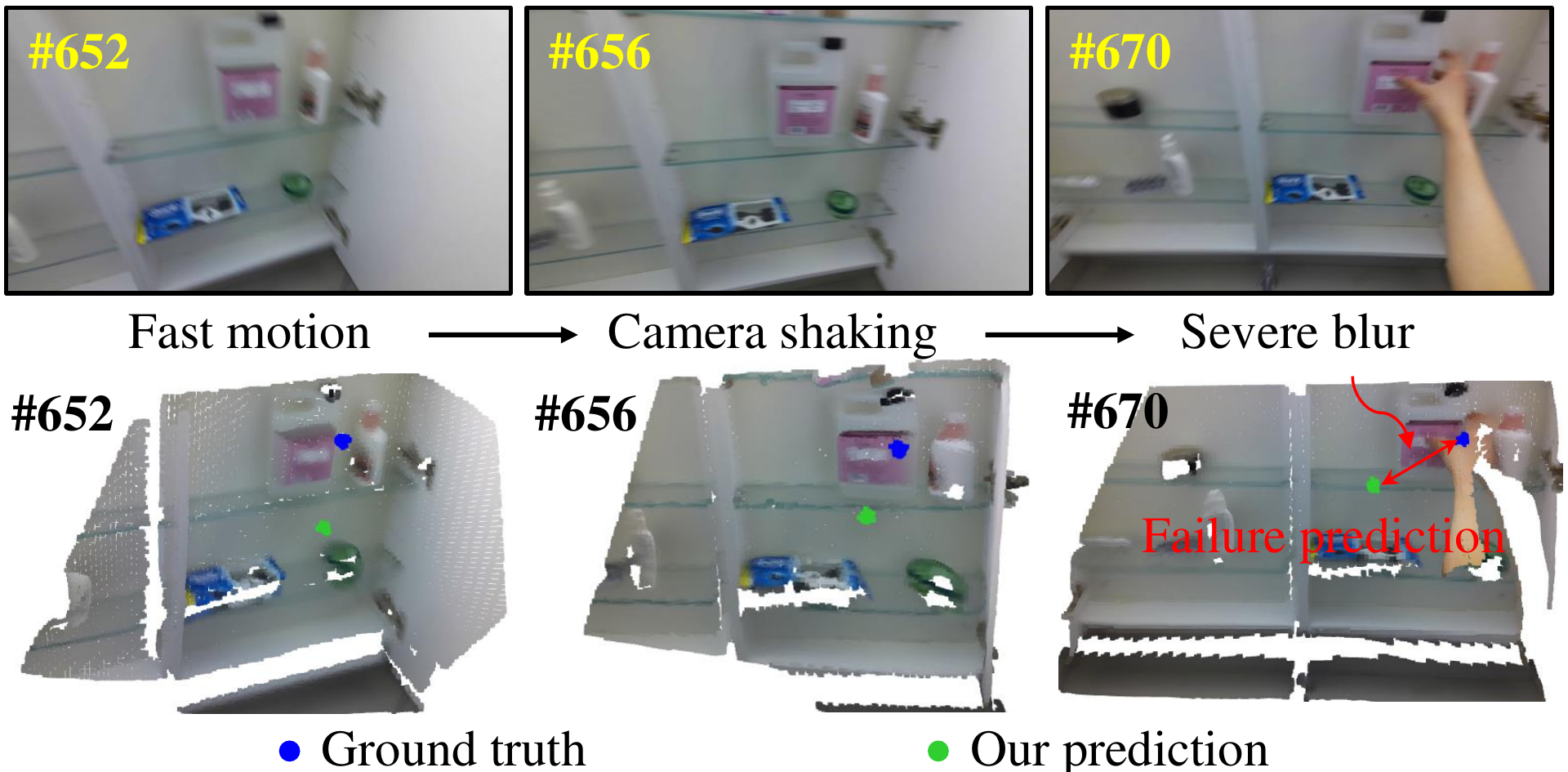}
  \caption{Visualization of one failure case.}
  \label{fig:fail}
\end{figure}

We will add failure case discussions in the final version. Here we show one challenging case with motion blur due to severe viewpoint change in Fig.~\ref{fig:fail}. In addition, we will clarify the discussion on motion features and provide analyses about our experiment results.

\section{Visualizations of evaluation results}
\begin{figure*}[t]
    \centering
    \includegraphics[width=\textwidth]{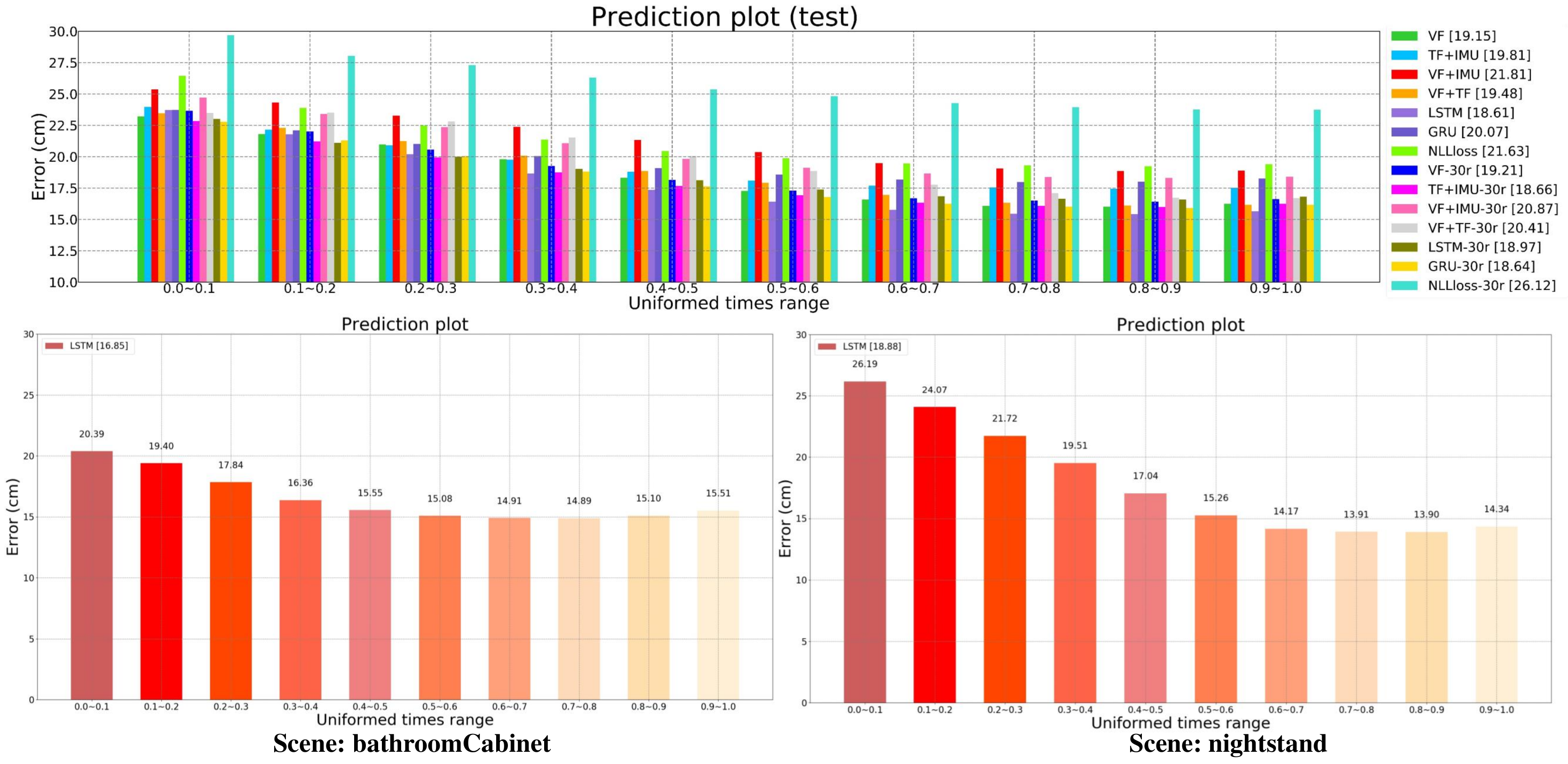}
    \includegraphics[width=\textwidth]{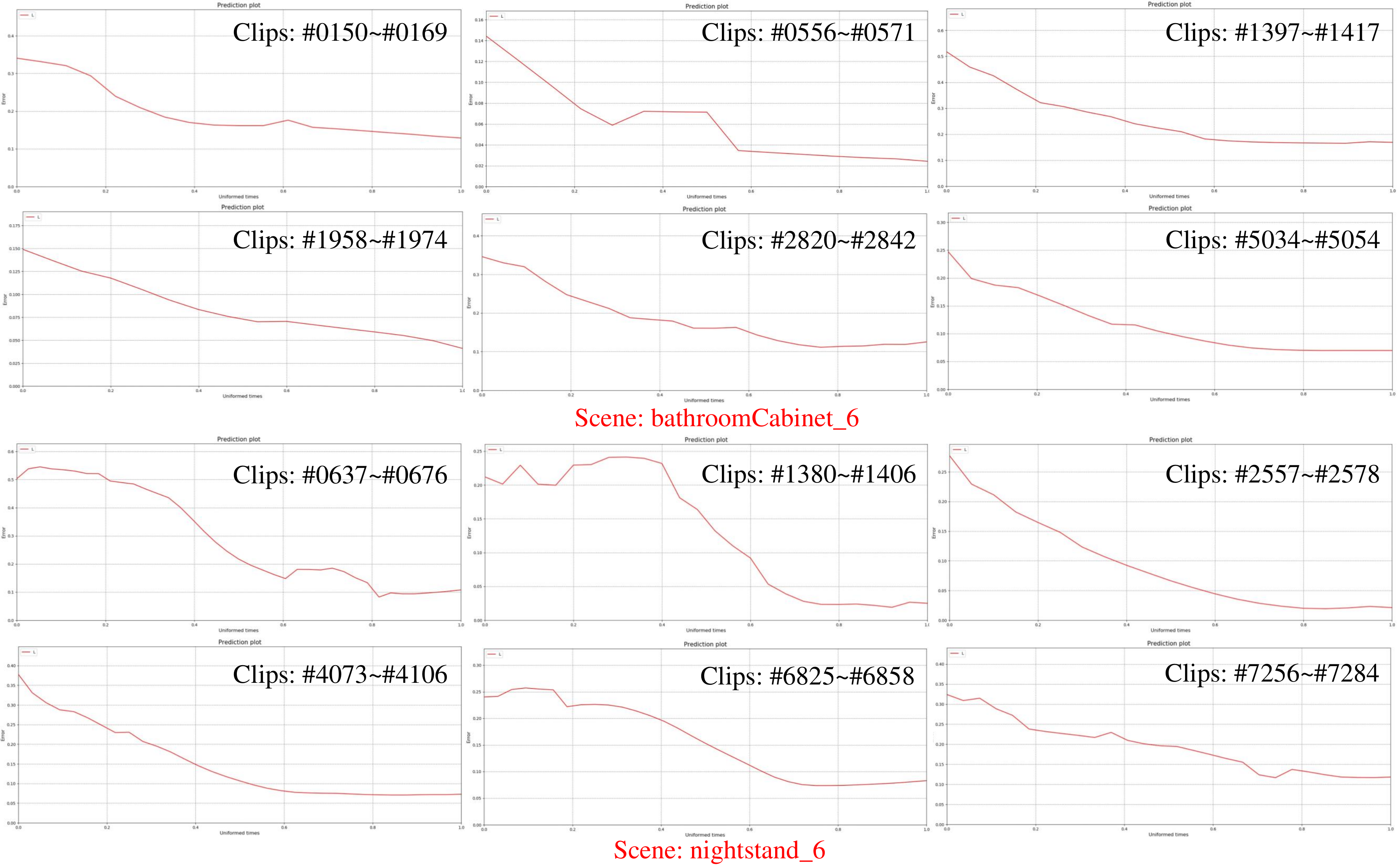}
    \caption{{Visualizations of errors in different stages.}}
    \label{fig:error}
\end{figure*}

The visualizations of errors in different time stages are shown in Fig.~\ref{fig:error}.

\end{document}